\documentclass[letterpaper]{article}
\usepackage[preprint]{aaai2027}
\usepackage[hyphens]{url}
\usepackage{graphicx}
\usepackage{natbib}
\usepackage{caption}
\usepackage{amsmath}
\usepackage{amssymb}
\usepackage{booktabs}
\usepackage{multirow}
\usepackage{makecell}
\usepackage{colortbl}
\usepackage{booktabs}
\usepackage{float}
\usepackage{microtype}
\usepackage{mwe}
\usepackage[most]{tcolorbox}
\usepackage{listings}

\definecolor{editgreen}{HTML}{1B6E3C}
\definecolor{editred}{HTML}{B22222}
\definecolor{editblue}{HTML}{1D4ED8}
\definecolor{lightblue}{HTML}{EAF2FF}
\definecolor{lightgreen}{HTML}{EAF7EF}
\newcommand{\cmark}{\textcolor{editgreen}{\ensuremath{\checkmark}}}
\newcommand{\xmark}{\textcolor{editred}{\ensuremath{\times}}}

\title{Edit2TikZ: A Comprehensive and Challenging Benchmark for Scientific Figure Editing with TikZ}
\author{
    Zongyun Zhang\textsuperscript{\rm 1}, 
    Jiacheng Ruan\textsuperscript{\rm 1}, 
    Xian Gao\textsuperscript{\rm 1}, 
    Ruizhu Zhou\textsuperscript{\rm 1}, \\
    Lingcheng Meng\textsuperscript{\rm 1}, 
    Lining Hu\textsuperscript{\rm 1}, 
    Ting Liu\textsuperscript{\rm 1}, 
    Yuzhuo Fu\textsuperscript{\rm 1}
}
\affiliations{
Shanghai Jiao Tong University\textsuperscript{\rm 1} \\
\texttt{zy.zhang2024@sjtu.edu.cn}
}

\begin{document}

\maketitle

\begin{abstract}
Although multimodal large language models (MLLMs) have shown substantial potential in visual understanding and graphic code generation, editing scientific figures through code presents a greater challenge: a model must jointly recover visual structure, ground the requested change, generate compilable code, and preserve all unrelated content. While existing TikZ benchmarks mainly focus on figure reconstruction and generation, few systematically evaluate instruction-guided scientific figure editing with compilable code. We introduce \textbf{Edit2TikZ}, a comprehensive benchmark for scientific figure editing tasks, featuring 1,548 diverse and high-quality samples. Edit2TikZ combines real-world and controlled synthetic edit cases, supports both textual and visual localization request, and contains multi-step editing, each with step-level annotations. We further construct a human-aligned evaluation framework to measure whether a requested edit is completed while irrelevant content is preserved. Utilizing Edit2TikZ, we evaluate 14 mainstream MLLMs and find that current systems remain unreliable: on average, proprietary models achieve a compilation success rate of merely 75\% and remain limited in both figure restoration and edit correctness, while compact models below 9B struggle further with instruction following and complete figure generation. Therefore, we build a mixed training set \textbf{TikZEditMix} and adopt reconstruction-then-editing curriculum learning for compact models. On Qwen3.5-4B, this training improves the compilation success rate from 45.35\% to 83.40\% and yields an average improvement of 18.7 points across our proposed evaluation metrics. The code and data will be released at \pdfstartlink attr{/Border[0 0 0]} user{/Subtype/Link/A<</S/URI/URI(https://github.com/Solunny/Edit2TikZ)>>}\url{https://github.com/Solunny/Edit2TikZ}\pdfendlink.
\end{abstract}

\section{Introduction}
\begin{figure}[t]
    \centering
    \includegraphics[width=0.995\columnwidth]{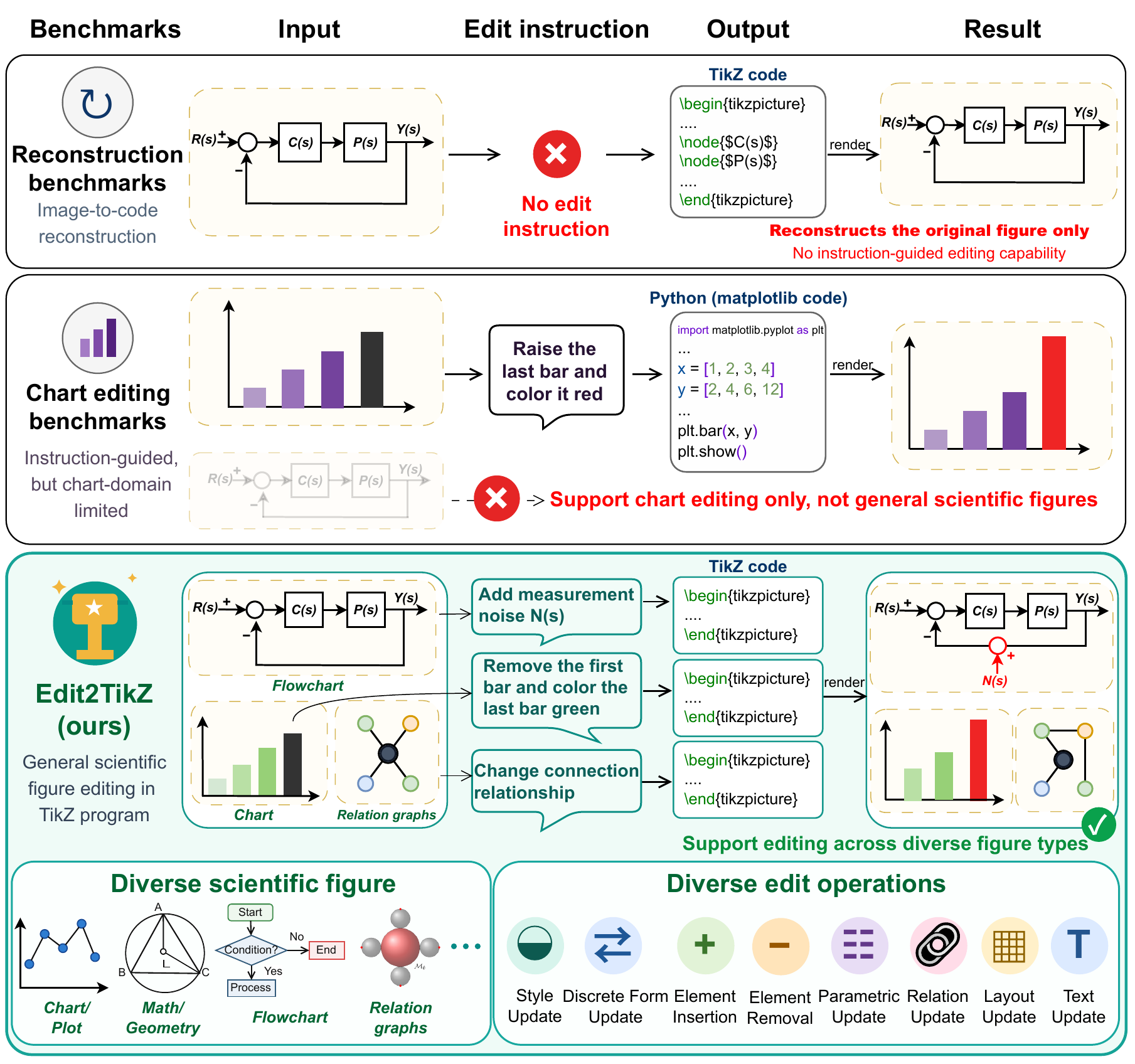}
    \caption{Previous benchmarks mainly focus on figure reconstruction or chart editing, whereas Edit2TikZ supports varied edit requests across a broad range of figure types.}
    \label{fig:motivation}
\end{figure}

Multimodal large language models (MLLMs) have shown substantial potential in visual understanding, reasoning, and code generation \cite{liu2023visual}. These capabilities create an opportunity for end-to-end scientific-figure editing: when source code is unavailable, a model may receive only a rendered figure and an edit request, then output a compilable graphics program that realizes the requested change. As illustrated in Figure~\ref{fig:motivation}, this task must support localized revisions across diverse figure types rather than a single visual genre. This workflow is substantially harder than recognizing figure content or generating code from text alone. The model must recover latent graphical structure from pixels, ground the objects and relations targeted by the instruction, produce a compilable program, and preserve unchanged content.

Existing benchmarks still leave this ability insufficiently evaluated. Graphics-program synthesis benchmarks cover the generation of programs from scientific figures, sketches, or text, including DeTikZify \cite{detikzify}, TikZero \cite{tikzero}, GeoTikzBridge \cite{geotikzbridge}, and broader image-to-code benchmarks \cite{vision2code}. However, these benchmarks mainly evaluate reconstruction or generation, and they do not support edit instructions that test whether a model can modify the requested content while preserving the rest of the figure. 

\begin{table*}[htbp]
    \centering
    \small
    \begin{tabular}{@{}p{0.235\textwidth}cclccp{0.22\textwidth}c@{}}
    \toprule
    Resource & \shortstack{Edit\\task} & \shortstack{Multi-step\\editing} & Source & \shortstack{w/ visual\\prompt} & \shortstack{Edit\\types} & Scope & Number \\
    \midrule
    DeTikZify \shortcite{detikzify} & \xmark & --  & Real & \xmark & -- & TikZ scientific figures & 1,000\\
    GeoTikZBridge \cite{geotikzbridge} & \xmark & --  & Synthetic & \xmark & -- & Geometry & 1,000\\
    vTikZ\cite{llmcode} & \cmark & \xmark  & Real & \xmark & 3 & TikZ scientific/animal figures & 100 \\
    ChartEdit\cite{chartedit} & \cmark & \xmark  & Both & \xmark & 6 & Chart & 1,405\\
    \rowcolor{lightgreen}\textbf{Edit2TikZ} & \cmark & \cmark &  \textbf{Both} & \cmark & \textbf{8 atomic} & \textbf{TikZ scientific figures} & \textbf{1,548}\\
    \bottomrule
    \end{tabular}
    \caption{Comparison of our proposed Edit2TikZ with other related graphics program generation benchmarks.}
    \label{tab:benchmark-comparison}
\end{table*}

Chart-editing benchmarks further advance the study of visually grounded figure editing, but they mainly target plotting-library graphics such as bar charts, line charts, and scatter plots \cite{chartedit,charteditor,charte3,charteditbench}. This scope does not cover many general scientific figures, such as geometric diagrams, circuit schematics, and model architectures. Existing benchmarks therefore either lack editable instructions, separate visual reconstruction from code editing, or focus on chart-style graphics. As a result, a dedicated benchmark for end-to-end editing of diverse scientific figures into compilable TikZ code is still urgently needed.

To address this gap, we introduce \textbf{Edit2TikZ}, a benchmark of 1,548 samples for scientific-figure editing. Each sample pairs a source figure with an editing instruction and requires the model to produce compilable TikZ code for the revised figure. The benchmark combines real-world and controlled synthetic edit cases and supports both text-based and visual localization requests. We define eight atomic edit operations, and each sample comprises one or more edit units drawn from them. All edit units are annotated and human-verified. We further construct a human-aligned evaluation framework to assess whether the requested modifications are completed while other text, objects, relations, and layout are preserved.

We conduct experiments of 6 open-source and 8 proprietary models on our Edit2TikZ. The results reveal that current systems remain unreliable: proprietary models compile and render only about 75\% of test examples on average and remain limited in both figure restoration and edit correctness, while compact models below 9B struggle further with instruction following and complete figure generation. These results show that end-to-end scientific-figure editing requires more than instruction understanding; it also depends on reliable image-to-TikZ reconstruction.

To strengthen compact models on this task, we construct \textbf{TikZEditMix}, a mixed training set that combines scientific-figure editing samples with image-to-TikZ reconstruction samples from DaTikZv3 \cite{tikzero}, and conduct a two-stage curriculum: first learning image-to-TikZ reconstruction, then specializing in scientific-figure editing. On Qwen3.5-4B, this training raises the compilation success rate from 45.35\% to 83.40\% and improves our proposed metrics by 18.7 points on average.

Our key contributions are as follows:
\begin{itemize}
    \item We introduce Edit2TikZ, a comprehensive benchmark for scientific-figure editing that spans diverse figure sources, edit requests, and edit operations.
    \item We construct a human-aligned evaluation framework that jointly assesses requested-edit completion and non-target preservation.
    \item Evaluation results on 14 MLLMs show that Edit2TikZ remains highly challenging. We therefore build a specialized training set, \textbf{TikZEditMix}, and show that two-stage curriculum learning substantially improves compact-model performance.
\end{itemize}

\section{Related Work}

\subsection{Graphics program generation.}
Existing TikZ benchmarks mainly focus on reconstruction or synthesis: DeTikZify reconstructs TikZ from scientific figures and sketches \cite{detikzify}, TikZero studies text-guided generation \cite{tikzero}, GeoTikzBridge extends image-to-TikZ supervision for geometry \cite{geotikzbridge}, and recent work further improves scientific graphics synthesis with reinforcement learning \cite{scientificgraphics}. Related image-to-code benchmarks also explore other target languages, including Matplotlib plots and charts \cite{plot2code,chartmimic,chartcoder}, HTML/CSS webpages \cite{design2code}, SVG graphics \cite{starvector}, and multi-domain compilable code generation \cite{vision2code}. These works substantially advance visual program generation, but they primarily evaluate reconstruction or synthesis rather than instruction-driven editing.

\subsection{Visually grounded editing.}
Recent editing benchmarks study how models revise structured visual outputs under user instructions. SVGEditBench V2 evaluates code-based SVG editing \cite{svgeditbenchv2}, and VCG-Bench unifies structured visual generation and editing within a broader benchmark framework \cite{vcgbench}. Chart-centered benchmarks further examine instruction-following visual editing, but their targets are mainly plotting-library graphics \cite{chartedit,chartm3,charteditbench,charte3,charteditor}. These benchmarks establish useful editing protocols, yet they either mainly study code-conditioned editing or focus on chart-style graphics.
\section{Edit2TikZ}

\subsection{Task Definition}

An Edit2TikZ sample is a tuple $(I_s,q^{\mathrm{text}},q^{\mathrm{vis}},y^*,I_t,E)$, where $I_s$ is the unmodified source figure, $q^{\mathrm{text}}$ is the edit instruction, $q^{\mathrm{vis}}$ is an optional visual localization prompt that highlights the target editing region, $y^*$ is the gold edited TikZ/TeX program, $I_t$ is its rendering, and $E$ is a sequence of image-verifiable edit units. For text-only cases, $q^{\mathrm{vis}}=\varnothing$; otherwise it comprises translucent red boxes overlaid on the source image. Let $\oplus$ denote this overlay operation. A model $p_\theta$ receives $(I_s\oplus q^{\mathrm{vis}},q^{\mathrm{text}})$ and generates a complete program $\hat y$:
\begin{equation}
    \hat y \sim p_\theta(y \mid I_s\oplus q^{\mathrm{vis}},q^{\mathrm{text}}).
\end{equation}

The edit sequence $E$ decomposes the instruction into localized steps, where each unit $e_i$ belongs to one of the predefined atomic operation types in $\mathcal{O}$. Thus,
\begin{equation}
    E=(e_1,\ldots,e_m), \qquad e_i\in\mathcal{O}.
\end{equation}

\paragraph{Atomic edits.}
We define $\mathcal{O}$ as eight atomic operations: (1) \textbf{appearance/style update}, modifying visual properties such as color, stroke, or fill; (2) \textbf{discrete form/representation update}, replacing an element's shape or visual representation; (3) \textbf{element insertion}, adding a new visible object; (4) \textbf{element removal}, deleting an existing visible object; (5) \textbf{metric/parametric update}, adjusting quantities such as size, angle, or length; (6) \textbf{reference/binding update}, changing the association or connection between elements; (7) \textbf{spatial arrangement/layering update}, modifying position, alignment, or occlusion order; and (8) \textbf{text/symbol content update}, revising visible labels, annotations, or mathematical symbols. Each edit unit describes the visible outcome in the target rendering rather than an implementation-specific TikZ command.

\subsection{Data Construction}
\label{sec:data-construction}

\subsubsection{Data Collection}
We collect scientific figures from recent arXiv TeX sources and extract self-contained TikZ code blocks with provenance metadata. Each block is compiled, rendered, and foreground-cropped through a shared pipeline, producing a source pool of programs, normalized renderings, and source URI. We use pHash to remove approximately duplicate samples that share the same source URI.\footnote{URI (uniform resource identifier); here it refers to the canonical, versioned arXiv source identifier.}.

\subsubsection{Editing Data Construction}

\begin{figure}[htbp]
    \centering
    \includegraphics[width=\columnwidth]{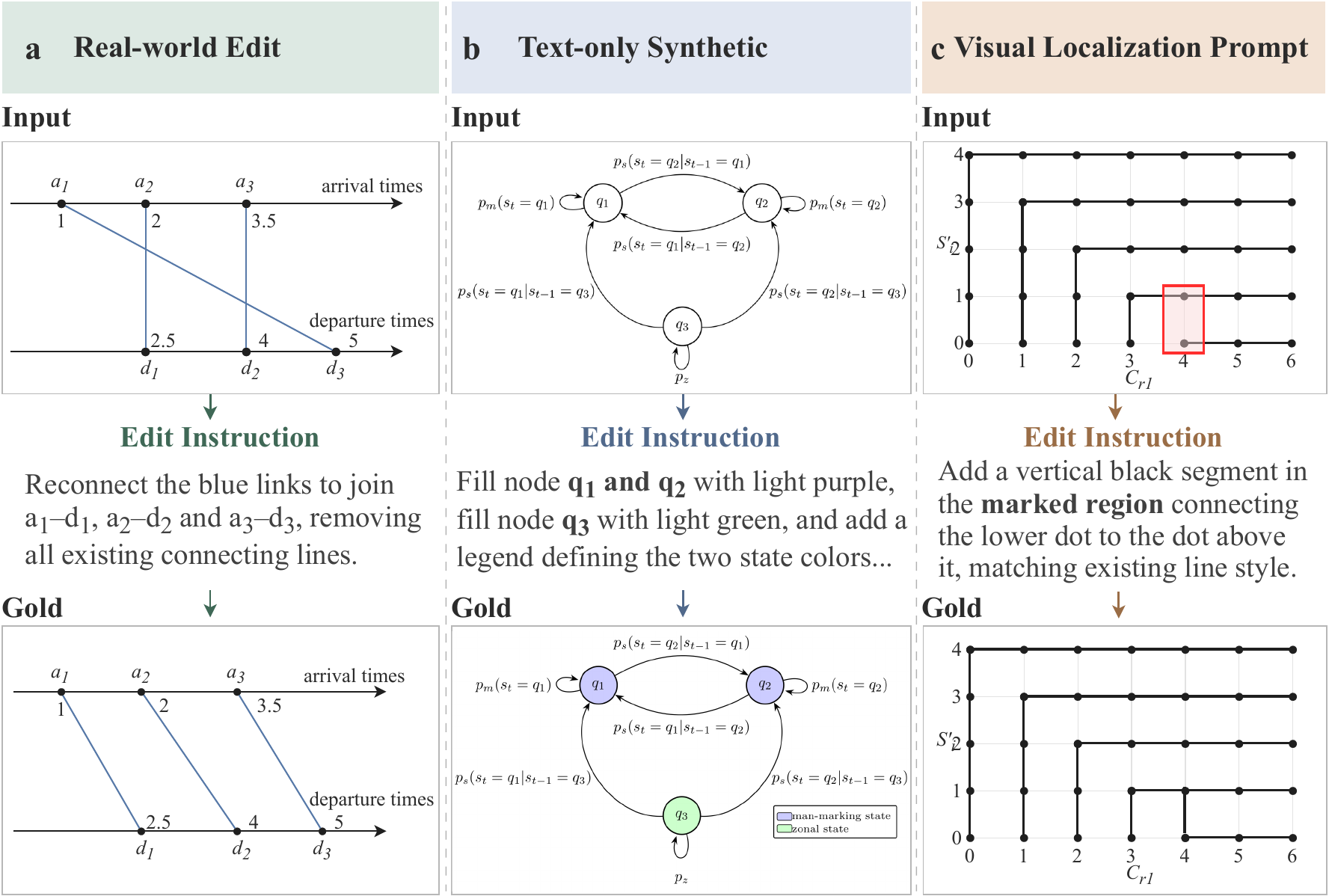}
    \caption{Examples from the real-world, text-only synthetic, and visual localization prompt synthetic subsets of Edit2TikZ.}
    \label{fig:construction-examples}
\end{figure}

\paragraph{Real-world edits.}
Figures within the same paper often describe related content, so they may share an overall structure while differing in key details such as labels, objects, relations, styles, or layouts. We use such variations to construct real-world editing pairs. Specifically, we identify candidate pairs using code similarity and foreground-aware visual similarity, and then rank and filter them to prevent repeated use of the same figure. For each retained pair, an instruction-generation model observes the source and target renderings, \(I_s\) and \(I_t\), together with their TikZ programs, and generates an instruction describing the required transformation. An edit-unit annotation model then decomposes the instruction into \(E=(e_1,\ldots,e_m)\) and assigns each edit unit \(e_i\) to one of the eight atomic operation types. Human reviewers verify the figure pair, instruction, edit-unit decomposition, and operation labels.

\paragraph{Controlled synthetic edits.}
Real-world samples are valuable but sparse and imbalanced, so we construct controlled synthetic edits from unused source figures.  Given a source rendering \(I_s\), its TikZ program \(x_s\), and the associated caption metadata, a planner produces a text instruction \(q^{\mathrm{text}}\) and edit units \(E\). An editing model applies the planned edits to \(x_s\), producing a candidate target program \(\tilde{y}\),which is compiled into a candidate target rendering \(\tilde{I}_t\). Deterministic checks and a multimodal validator then assess compilability, instruction satisfaction, and non-target preservation. Failed candidates receive validator feedback and undergo at most two repair attempts, while all automatically accepted candidates remain subject to human review. The accepted program and its rendering are denoted by \(y^*\) and \(I_t\), respectively.

\paragraph{Visual localization prompt.}
Scientific figures often contain repeated, unlabeled, or visually similar elements that cannot be identified reliably through language alone. We therefore provide visual localization prompt \(q^{\mathrm{vis}}\) that directly highlight the regions to be edited. When a text instruction cannot distinguish nearby elements unambiguously, the planner jointly produces \(q^{\mathrm{text}}\), \(E\), and one or more target-region boxes. These boxes are overlaid on \(I_s\), yielding the visually prompted input \(I_s \oplus q^{\mathrm{vis}}\). Figure~\ref{fig:construction-examples} presents representative samples from the three resulting subsets.

\begin{figure}[htbp]
    \centering
    \includegraphics[width=0.97\columnwidth]{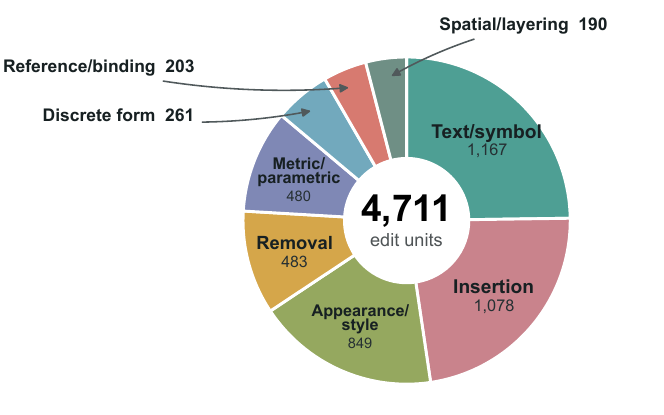}
    \caption{Atomic-operation coverage of our benchmark. In Edit2TikZ, 1,548 samples contain 4,711 labeled edit units covering all eight operation types.}
    \label{fig:benchmark-distribution}
\end{figure}

\subsection{Data Filtering}
Human reviewers inspect every sample, checking the source and target renderings, instruction, visual localization boxes and edit units. They reject samples with imperceptible edits, damage to non-target content, ambiguous instructions, or inaccurate localization, and convert unnecessary visual prompt into text-only instructions. The final selection balances real-world, text-only synthetic, and visual prompt synthetic edits across program lengths, edit-unit counts, and the eight atomic operations.

\subsection{Data Statistics}
The final Edit2TikZ benchmark contains 534 real-world, 580 text-only synthetic, and 434 visual prompt synthetic samples. Each instruction comprises one to six labeled edit units (3.04 on average): 689 samples contain one or two units, 546 contain three or four, and 313 contain five or six. As shown in Figure~\ref{fig:benchmark-distribution}, the 1,548 samples comprise 4,711 edit units spanning all eight atomic operations.

\subsection{Evaluation Metrics}
\label{sec:benchmark-evaluation}
For comparability with prior TikZ reconstruction evaluations, we report compilation success rate, \textbf{cBLEU}, and \textbf{TED}, with the latter two measuring token-level overlap and structural distance between predicted and gold programs \cite{detikzify}. We also report \textbf{DSim}, computed as one minus the DreamSim perceptual distance between predicted and gold renderings \cite{dreamsim}. However, these metrics cannot explicitly assess whether the requested edits are completed while unedited content is preserved \cite{gao2026evaluating}.

To address this limitation, we construct a human-aligned evaluation framework with two complementary scores. \textbf{RestorationScore (RS)} measures non-target preservation on a 0--100 scale, allocating 30 points to text/style, 30 to objects/relations, and 40 to layout/scale. A primary-error rule prevents the same underlying failure from being penalized across multiple dimensions. \textbf{EditCorrectnessScore (ECS)} evaluates each labeled edit unit using six score levels: \(\{0,40,80,90,95,100\}\). A score of 0 indicates no completion, 40 partial completion, and 80 essential completion, though the result may not be perfect. Scores of 90 and 95 indicate progressively better layout coherence and visual quality, while 100 denotes a complete and perfect result. The sample-level ECS is the mean score across all edit units. Both RS and ECS are set to zero for rendering failures.

\paragraph{Human alignment.}
To assess the reliability and fairness of our evaluation framework, two human annotators and three candidate model judges independently score RS and ECS on 100 randomly sampled cases. Table~\ref{tab:alignment} reports human inter-rater reliability and human--AI agreement using mean absolute error (MAE), Pearson correlation \(r\), and Spearman rank correlation \(\rho\). 
Human--human agreement is measured between Human Annotators A and B, whereas human--AI agreement for each model judge is computed against the per-sample mean of the two human ratings. Both RS and ECS exhibit strong human consensus and human--AI alignment, with all reported correlation coefficients exceeding 0.70. Among the candidate judges, GPT-5.6-Terra achieves the lowest MAE for both metrics while maintaining consistently high Pearson and Spearman correlations. We therefore select it as the final judge model. In contrast, DSim yields Pearson correlations of only 0.496 with human RS and 0.230 with human ECS, indicating that global image similarity cannot reliably evaluate our figrue editing task.

\begin{table}[htbp]
    \centering
    \small
    \setlength{\tabcolsep}{0.25mm}
    \begin{tabular}{@{}cccccc@{}}
    \toprule
    Metrics&\shortstack{Comparison} & Models & MAE $\downarrow$ & \shortstack{Pearson \\ $r$ $\uparrow$} & \shortstack{Spearman \\ $\rho$  $\uparrow$} \\
    \midrule
    \multirow{4}{*}{\textbf{RS}}&Human-Human & - & 19.01 & 0.741 & 0.771 \\
    \cmidrule{2-6}
    &Human-AI & GPT-5.6-Terra & \textbf{10.19} & 0.763 & 0.700 \\
    &Human-AI & GPT-5.6-Sol & 10.67 & \textbf{0.781} & \textbf{0.750} \\
    &Human-AI & GPT-5.6-Luna & 11.29 & 0.744 & 0.712 \\
    \midrule
    \multirow{4}{*}{\textbf{ECS}} &Human-Human & - & 16.18 & 0.770 & 0.765 \\
    \cmidrule{2-6}
    &Human-AI & GPT-5.6-Terra & \textbf{12.82} & \textbf{0.841} & 0.800 \\
    &Human-AI & GPT-5.6-Sol & 13.16 & 0.835 & \textbf{0.819} \\
    &Human-AI & GPT-5.6-Luna & 16.87 & 0.767 & 0.743 \\
    \bottomrule
    \end{tabular}
    \caption{Human--AI agreement and human inter-rater reliability for RS and ECS.}
    \label{tab:alignment}
\end{table}

\begin{table*}[t]
    \centering
    \small
    \setlength{\tabcolsep}{1.0mm}
    \begin{tabular}{@{}lcccccccccr@{}}
    \toprule
    & & & & & \multicolumn{4}{c}{RS $\uparrow$} & & \\
    \cmidrule(lr){6-9}
    Model & Comp. (\%) $\uparrow$ & cBLEU$^{\dagger}$ $\uparrow$ & TED$^{\dagger}$ $\downarrow$ & DSim $\uparrow$ & \shortstack{Text/Style\\(30)} & \shortstack{Objects/\\Relations (30)} & \shortstack{Layout/Scale\\(40)} & \shortstack{Total\\(100)} & ECS(100) $\uparrow$ & \shortstack{Code\\lines} \\
    \midrule
    \rowcolor{lightblue}\multicolumn{11}{c}{\textbf{Proprietary Models}} \\
    \midrule
    qwen3.7-plus & 67.05 & 16.44 & 49.88 & 0.58 & 16.99 & 16.41 & 16.52 & 49.93 & 49.04 & 71 \\
    Doubao-Seed-2.1-Pro & 65.76 & 15.54 & 50.45 & 0.55 & 16.41 & 15.96 & 15.57 & 47.94 & 49.88 & 61 \\
    GPT-5.6-Luna & 74.10 & 13.65 & 50.19 & 0.64 & 19.18 & 18.82 & 18.27 & 56.26 & 54.79 & 66 \\
    GPT-5.6-Terra & 77.33 & 14.68 & 49.96 & 0.69 & 20.48 & 21.12 & 20.69 & 62.29 & 61.04 & 68 \\
    GPT-5.6-Sol & 75.65 & 14.91 & 50.27 & 0.68 & 20.44 & 21.42 & 21.38 & 63.24 & 64.10 & 80 \\
    Claude-Sonnet-4.6 & 74.10 & 16.86 & 50.25 & 0.64 & 17.87 & 17.81 & 18.17 & 53.85 & 59.55 & 92 \\
    Claude-Opus-4.8 & 78.88 & 15.54 & 50.21 & 0.69 & 20.36 & 21.10 & 20.63 & 62.09 & 64.98 & 65 \\
    Gemini-3.1-Pro & \textbf{88.37} & \textbf{18.64} & \textbf{48.12} & \textbf{0.81} & \textbf{24.04} & \textbf{24.73} & \textbf{25.74} & \textbf{74.51} & \textbf{75.02} & 57 \\
    \midrule
    \rowcolor{lightgreen}\multicolumn{11}{c}{\textbf{Open-Source Models}} \\
    \midrule
    Qwen3.5-2B & 13.24 & 1.28 & 75.29 & 0.09 & 2.30 & 1.77 & 2.02 & 6.09 & 2.78 & 422 \\
    Qwen3.5-4B & 45.35 & 7.83 & 56.02 & 0.34 & 9.50 & 8.01 & 8.59 & 26.10 & 22.54 & 124 \\
    Gemma-4-E2B-it & 6.72 & 10.45 & 55.06 & 0.05 & 1.12 & 1.06 & 0.98 & 3.17 & 2.97 & 91 \\
    Gemma-4-E4B-it & 52.39 & 11.70 & 53.14 & 0.38 & 8.82 & 7.50 & 8.32 & 24.64 & 30.18 & 82 \\
    Qwen3.5-9B & 52.26 & 7.71 & 56.26 & 0.41 & 11.50 & 10.06 & 10.79 & 32.35 & 24.99 & 121 \\
    Qwen3.6-27B & 69.77 & 16.12 & 50.77 & 0.58 & 16.32 & 15.84 & 16.41 & 48.57 & 48.50 & 83 \\
    \bottomrule
    \end{tabular}
    \caption{Performance of 14 mainstream MLLMs on our Edit2TikZ benchmark. Performance is evaluated using compilation success rate (Comp.), code-level metrics (cBLEU and TED), image-based metrics (DSim, RS, and ECS), and the average number of generated code lines. $^{\dagger}$ indicates that both two code-level metrics are reported on a $0,100$ scale and computed over all 1,548 test samples.}
    \label{tab:general-results}
\end{table*}

\section{Experiments}
\label{sec:experiments}

\subsection{Baselines and Evaluation Metrics}

To effectively compare the capabilities of existing MLLMs on the Edit2TikZ task, we evaluate 14 mainstream multimodal models. These include 8 proprietary models: Qwen3.7-plus \cite{qwen35omni}, Doubao-Seed-2.1-Pro \cite{seed20}, GPT-5.6 Luna, Terra, and Sol \cite{openai_gpt5}, Claude Sonnet 4.6, Claude Opus 4.8 \cite{anthropic_claude4}, and Gemini-3.1-Pro-Preview \cite{gemini3procard}. We also evaluate 6 open-source models: Gemma-4-E2B-it, Gemma-4-E4B-it \cite{gemma4}, Qwen3.5-2B, Qwen3.5-4B, Qwen3.5-9B, and Qwen3.6-27B \cite{qwen35omni}. 

We evaluate each model using the compilation success rate, the code-level metrics cBLEU and TED following DeTikZify~\cite{detikzify}, and the image-based metrics DSim, RS, and ECS defined in Section~\ref{sec:benchmark-evaluation}. We also report the mean number of code lines generated by each model. All outputs are compiled in a shared TeX Live environment, rendered, foreground-cropped, and resized to a maximum long-edge length of 1,024 pixels. Rendering failures receive zero for DSim, RS, and ECS.

\subsection{Main Results}
The experimental results on Edit2TikZ are summarized in Table~\ref{tab:general-results}. 
Overall, the eight proprietary models achieve average scores of 75.15\% compilation success, 58.76 RS, and 59.80 ECS, whereas the six open-source models reach only 39.95\%, 23.49, and 21.99, respectively. Even the strongest proprietary model, Gemini-3.1-Pro, still fails on about 12\% of the test samples, showing that current proprietary models are still far from reliable end-to-end scientific-figure editing. Open-source scaling does improve performance within model families; for example, within Qwen, moving from 2B to 27B raises compilation success from 13.24\% to 69.77\%, RS from 6.09 to 48.57, and ECS from 2.78 to 48.50. However, this task remains highly challenging for open models even at larger scales. Beyond these overall trends, we highlight four additional findings below.

\paragraph{(1) Reference-code similarity does not adequately reflect editing quality.} Among proprietary models, cBLEU and TED vary within relatively narrow ranges (13.65--18.64 and 48.12--50.45), while ECS spans 49.04--75.02. The same mismatch appears among open-source models: Gemma-4-E2B-it and Gemma-4-E4B-it obtain similar cBLEU scores (10.45 vs.\ 11.70), yet their ECS differs by 27.21 points (2.97 vs.\ 30.18). This weak correspondence suggests that code similarity to the gold program cannot reliably capture functionally equivalent TikZ implementations or determine whether the requested edit was completed.

\paragraph{(2) Program completeness is a primary failure mode for compact models.} Models with 9B parameters or fewer frequently produce repetitive outputs, even under stronger repetition penalties. One reason is that TikZ programs naturally contain recurring drawing commands, environments, style declarations, and coordinate patterns, which may trigger generation loops. Another is that end-to-end editing requires the model to reconstruct the figure, execute localized edits, and maintain long-range TeX structure within a one time generation, making compact models prone to generation loops and failure to emit \texttt{\textbackslash end\{document\}}. Among compilation failures from Qwen3.5-2B, 4B, and 9B, 77.1\%, 29.4\%, and 36.5\% have incomplete document structures. Their failed outputs average 479, 176, and 194 lines, versus 45, 62, and 54 lines for successful outputs. Thus, low compilation success reflects both limited TikZ knowledge and repetition or termination failures in long-program generation.

\paragraph{(3) Better reconstruction does not automatically yield better editing.} Scaling Qwen3.5 from 4B to 9B raises all-example RS from 26.10 to 32.35, but ECS increases only from 22.54 to 24.99. To control for compilation, we further compare the 461 samples rendered successfully by both models. The 9B model raises mean RS from 60.74 to 65.09, yet its mean ECS slightly decreases from 50.23 to 49.22. This paired trend shows that more complete source-figure reconstruction does not ensure completion of the requested modifications: visual reconstruction and localized instruction following remain separable capabilities.

\paragraph{(4) Difficulty varies systematically with program structure and edit type.} Model performance declines as both source-program length and instruction complexity increase. Averaged across all 14 models, compared with 1--50-line source programs and instructions containing one or two edit units, programs longer than 100 lines and instructions containing five or six edit units raise the failure rate by 19.13 and 12.68 percentage points, respectively. 

\begin{figure}[htbp]
    \centering
    \includegraphics[width=0.95\columnwidth]{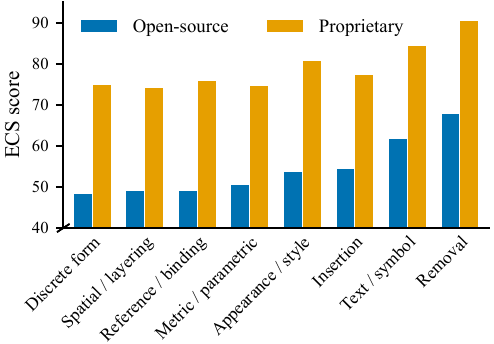}
    \caption{Operation-level ECS for the six open-source and eight proprietary models. Each ECS value pools unit scores from successfully rendered outputs across every model in the corresponding group.}
    \label{fig:operation-difficulty}
\end{figure}

Figure~\ref{fig:operation-difficulty} aggregates the completed evaluations of all 14 models at the edit-unit level. It shows that discrete-form, spatial/layering, reference/binding, and metric/parametric updates are consistently the most difficult operations. Both model groups score relatively poorly on the hardest categories: for discrete-form and spatial/layering edits, open-source ECS is 48.31 and 49.07, compared with 74.90 and 74.27 for proprietary models, leaving gaps of 26.59 and 25.20 points. These operations often require coordinated changes to representations, coordinates, anchors, scopes, drawing order, relation endpoints, or coupled parameters rather than isolated parameter edits. By contrast, scores rise substantially for the simpler text/symbol and removal operations, reaching 61.72 and 67.86 for open-source models and 84.46 and 90.53 for proprietary models. The corresponding gaps narrow to 22.74 and 22.67 points, suggesting that simpler localized edits reduce, but do not eliminate, the capability gap.

\subsection{Discussion}
\subsubsection{Evaluating Reconstruct-then-Edit}

Motivated by prior benchmarks that decompose this end-to-end editing task into vision-to-code generation and code-to-code editing~\cite{vcgbench}, we evaluate a two-step Recon.$\to$Edit protocol. The model first reconstructs a source program $\hat{y}_s$ from $I_s$, and then produces the edited program $\hat{y}_e$ from $I_s$, $\hat{y}_s$, and the edit request. To assess how source-program quality affects editing, we also evaluate Gold Source$\to$Edit, where the model produces $\hat{y}_e$ from $I_s$, the dataset-provided source program $y_s$, and the same edit request. Comparing the two protocols reveals the performance gap between editing from reconstructed and gold source programs.

\begin{table}[htbp]
    \centering
    \small
    \setlength{\tabcolsep}{1.0mm}
    \begin{tabular}{@{}llccc@{}}
    \toprule
    Model & Protocol & Comp.\% $\uparrow$ & RS $\uparrow$ & ECS $\uparrow$ \\
    \midrule
    \multirow{3}{*}{Qwen3.5-4B}  & Recon. & 54.01 & 31.08 & -- \\
    & Recon.$\to$Edit & 46.83 & 26.06 & 21.72 \\
    & Gold Source$\to$Edit & 64.66 & 60.47 & 36.76 \\
    \midrule
    \multirow{3}{*}{Qwen3.5-9B} & Recon. & 62.47 & 37.49 & -- \\
    & Recon.$\to$Edit & 60.34 & 36.70 & 30.13 \\
    & Gold Source$\to$Edit & 90.70 & 86.51 & 54.08 \\
    \midrule
    \multirow{3}{*}{Qwen3.6-27B} & Recon. & 75.26 & 50.64 & -- \\
    & Recon.$\to$Edit & 65.31 & 44.25 & 46.81 \\
    & Gold Source$\to$Edit & 90.76 & 86.87 & 74.04 \\
    \bottomrule
    \end{tabular}
    \caption{Performance under reconstruction-based inference protocols. Recon. evaluates source-program reconstruction from $I_s$; Recon.$\to$Edit performs editing with the reconstructed source program; and Gold Source$\to$Edit performs editing with the gold pre-edit program $y_s$. RS for Recon. measures reconstruction quality, whereas RS and ECS for the editing protocols follow our standard evaluation.}
    \label{tab:inference-protocols}
\end{table}

Compared with the results in Table~\ref{tab:general-results}, reconstruct-then-edit provides no consistent gain: compilation success changes by \(+1.48\), \(+8.08\), and \(-4.46\) percentage points for 4B, 9B, and 27B, respectively; RS changes by \(-0.04\), \(+4.35\), and \(-4.32\) points; and ECS changes by \(-0.82\), \(+5.14\), and \(-1.69\) points. In contrast, when the reconstructed source program provided to the editing stage is replaced with the gold source program, compilation success increases by \(17.83\), \(30.36\), and \(25.45\) percentage points for 4B, 9B, and 27B, respectively; RS increases by \(34.41\), \(49.81\), and \(42.62\) points, and ECS by \(15.04\), \(23.95\), and \(27.23\) points. These results identify source reconstruction as the main bottleneck: errors in $\hat{y}_s$ propagate to editing, whereas accurate source code enables substantially stronger edits.

\subsubsection{Agentic Iterative Refinement}
\begin{figure}[htbp]
    \centering
    \includegraphics[width=\columnwidth]{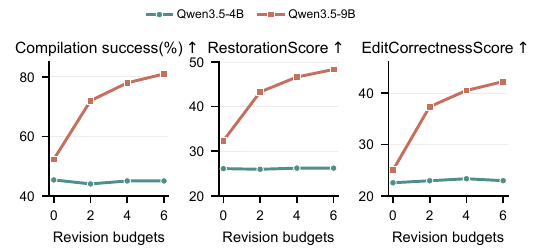}
    \caption{Compilation success rate, RS, and ECS under increasing revision budgets with compiler-and-renderer feedback. A budget of zero corresponds to a single generation without revision.}
    \label{fig:iterative-agent}
\end{figure}

We evaluate an iterative tool-assisted generation protocol in which a LaTeX compiler and renderer provide direct feedback on both program validity and visual outcome. For each candidate program, successful compilation returns the rendered image, while failure returns the compiler error. Based on this feedback, the model decides whether to terminate with \texttt{FINAL} or revise the program. We allow up to six revisions. Figure~\ref{fig:iterative-agent} reports compilation success and all-example semantic scores across revision budgets.

Iterative feedback yields clear gains for Qwen3.5-9B: compilation success, RS, and ECS improve sharply within the first two revisions, followed by smaller gains as the budget increases. In contrast, Qwen3.5-4B shows little improvement across all metrics, suggesting that effectively using compilation and visual feedback requires sufficient model capability rather than merely a larger revision budget.

\begin{figure*}[t]
    \centering
    \includegraphics[width=0.98\textwidth]{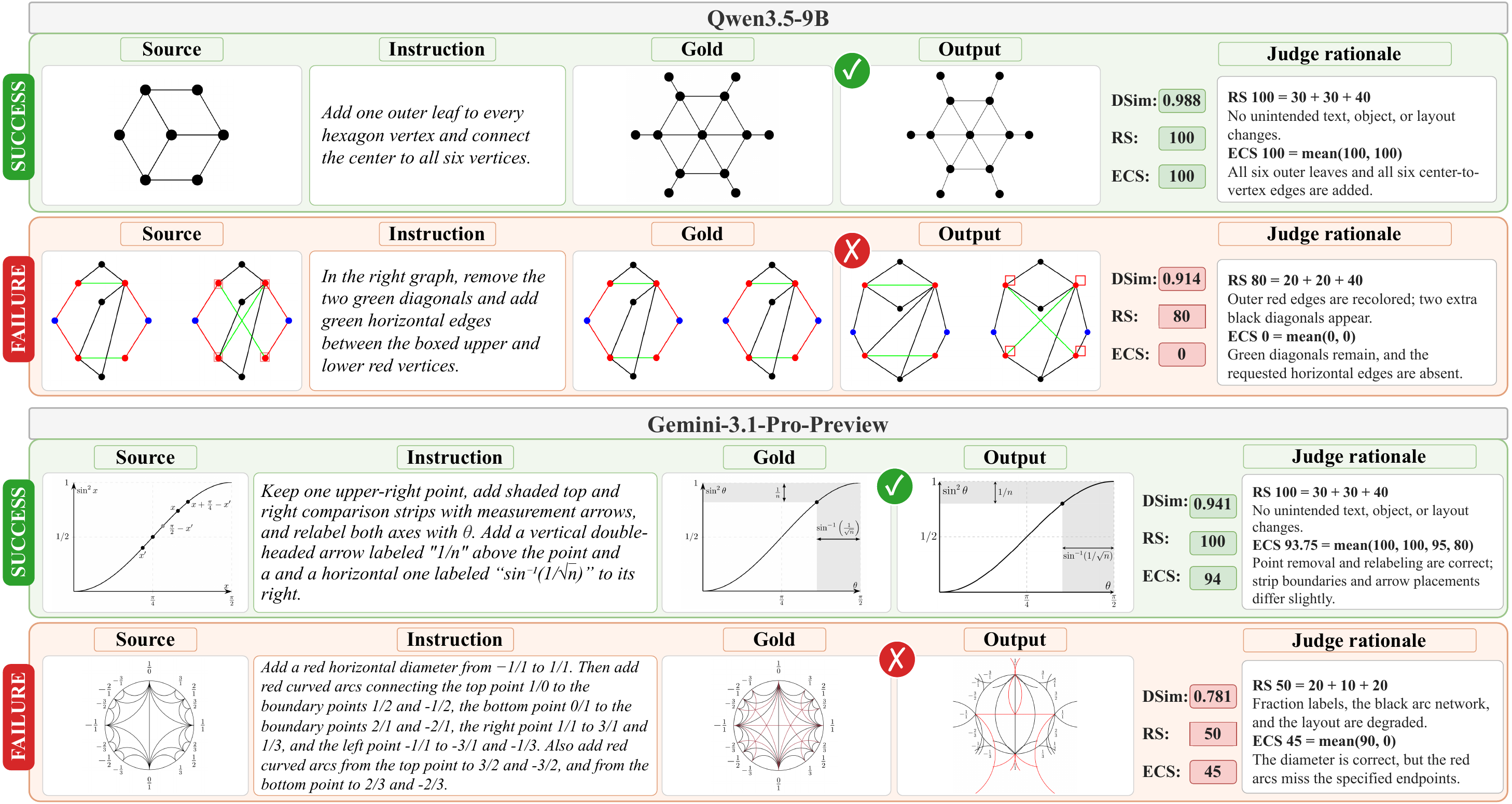}
    \caption{Qualitative examples of successful and failed cases. Models perform well on relatively simple figures with few elements and regular edit patterns, but struggle when edits involve dense relational structure, multiple subfigures, or precise preservation of existing geometry. RS and ECS are reported with judge rationales.}
    \label{fig:case-study}
\end{figure*}

\subsection{Training}

\subsubsection{Method and Dataset}
Scientific-figure editing requires recovering the complete source structure while applying localized modifications. We therefore construct \textbf{TikZEditMix}, comprising 32,448 samples, and adopt a two-stage training curriculum. The reconstruction subset contains 20,244 high-quality image-to-TikZ pairs filtered from DaTikZ-v3 \cite{tikzero}, while the editing subset contains 12,204 constructed scientific-figure editing samples. Stage~1 learns image-to-TikZ reconstruction from a rendering $I$ and its corresponding program $y$. Stage~2 learns instruction-conditioned editing from a source image $I_s$ and edit instruction $q$, supervised by the complete edited program $y^*$. The corresponding token-level objectives are
\begin{align}
    \mathcal{L}_{\mathrm{rec}}(\theta)
    &=-\sum_t \log p_\theta(y_t\mid I,y_{<t}),\\
    \mathcal{L}_{\mathrm{edit}}(\theta)
    &=-\sum_t \log p_\theta(y^*_t\mid I_s,q,y^*_{<t}).
\end{align}
The training data have no identifier or source-URI overlap with the 1,548-sample test set. We train Qwen3.5-2B and Qwen3.5-4B separately using this two-stage curriculum.

\begin{table}[htbp]
    \centering
    \small
    \setlength{\tabcolsep}{0.5mm}
    \begin{tabular}{@{}clcccc@{}}
    \toprule
    Model & Training phase & \shortstack{Comp.(\%)$\uparrow$} & DSim $\uparrow$ & RS $\uparrow$ & ECS $\uparrow$ \\
    \midrule
    \multicolumn{6}{c}{\textit{Equal-data single-stage ablation}} \\
    \cmidrule(lr){1-6}
    2B & Single-stage training & 63.31 & 0.47 & 31.47 & 18.98 \\
    4B & Single-stage training & 75.84 & 0.61 & 45.67 & 36.33 \\
    \midrule
    \multicolumn{6}{c}{\textit{Two-stage curriculum}} \\
    \cmidrule(lr){1-6}
    2B & Stage 1: Recon. & 67.38 & 0.48 & 30.61 & 24.52 \\
    \rowcolor{lightgreen}2B & \textbf{+Stage 2: Edit} & \textbf{69.77} & \textbf{0.53} & \textbf{35.78} & \textbf{29.53} \\
    \midrule
    4B & Stage 1: Recon. & 51.42 & 0.39 & 31.08 & 22.82 \\
    \rowcolor{lightgreen}4B & \textbf{+Stage 2: Edit} & \textbf{83.40} & \textbf{0.67} & \textbf{48.72} & \textbf{37.32} \\
    \bottomrule
    \end{tabular}
    \caption{Two-stage curriculum results and the equal-data single-stage ablation on Qwen3.5-2B and Qwen3.5-4B.}
    \label{tab:training-results}
\end{table}

\subsubsection{Experimental Setup}

Training begins with one epoch on the reconstruction subset at a learning rate of $10^{-5}$, followed by one epoch on the editing subset initialized from the Stage~1 checkpoint at a learning rate of $5\times10^{-6}$. To verify the advantage of curriculum learning over single-stage training, we conduct an ablation that mixes the reconstruction and editing subsets and jointly trains on the same 32,448 samples. All experiments are conducted on 8$\times$A800 GPUs.

\subsubsection{Results}

Table~\ref{tab:training-results} shows that the two-stage curriculum substantially improves over performance before task-specific training: Comp., RS, and ECS increase by 56.53, 29.69, and 26.75 points for 2B, and by 38.05, 22.62, and 14.78 points for 4B, respectively. Under the equal-data setting, it also outperforms single-stage training across all metrics, indicating that the gains arise from the staged curriculum rather than additional data. From Stage~1 to Stage~2, however, 2B gains only 2.39 points in Comp. and 5.01 in ECS, whereas 4B gains 31.98 and 14.50 points, suggesting that the smaller model approaches its capacity limit earlier, while the larger model better benefits from editing supervision.

\paragraph{External reconstruction benchmark.}
Beyond Edit2TikZ, we evaluate general reconstruction capability on the 442 publicly released DaTikZv2 test examples using the same rendering and evaluation protocol for every model. Table~\ref{tab:datikzv2-reconstruction} shows that our two-stage-trained Qwen3.5-4B achieves the best cBLEU and TED among the compared models despite using a 32k-sample training set, substantially smaller than the approximately 360K examples used to fine-tune the four DeTikZify models. Relative to its base checkpoint, training improves cBLEU from 2.945 to 4.215, reduces TED from 57.025 to 56.193, and raises DSim from 68.168 to 73.189. It also surpasses the base Qwen3.5-9B on all three metrics.

\begin{table}[htbp]
    \centering
    \begin{tabular}{@{}lccc@{}}
    \toprule
    Model & cBLEU $\uparrow$ & TED $\downarrow$ & DSim $\uparrow$ \\
    \midrule
    DT-TL1.1b & 1.715 & 58.914 & 67.803 \\
    DT-DS1.3b & 2.053 & 58.295 & 70.544 \\
    DT-CL7b & 2.263 & 57.285 & 74.098 \\
    DT-DS7b & 2.731 & 57.293 & \textbf{74.274} \\
    \midrule
    Qwen3.5-4B & 2.945 & 57.025 & 68.168 \\
    Qwen3.5-9B & 3.716 & 56.872 & 72.466 \\
    Qwen3.5-4B (Ours) & \textbf{4.215} & \textbf{56.193} & 73.189 \\
    \bottomrule
    \end{tabular}
    \caption{Reconstruction comparison of our two-stage-trained Qwen3.5-4B against its base checkpoint, the larger base Qwen3.5-9B, and four DeTikZify models on the 442 publicly released DaTikZv2 test examples. All results are obtained under the same evaluation protocol.}
    \label{tab:datikzv2-reconstruction}
\end{table}

\subsection{Qualitative Cases}
Figure~\ref{fig:case-study} presents representative successful and failed editing cases on Qwen3.5-9B and Gemini-3.1-Pro-Preview. These examples show that DSim alone may overestimate edit quality: in the failed Qwen3.5-9B case, the output remains globally similar to the target and achieves a DSim of 0.914, despite retaining edges that should be removed, omitting requested edges, and introducing unintended changes, resulting in an ECS of 0. The cases also indicate that the models handle simple and regular modifications more reliably than edits involving dense, interdependent point--edge relations, where precise endpoint correspondence becomes difficult. Moreover, failures often affect not only the requested modification but also preserved content; for example, Gemini distorts the original black arc structure while adding incomplete red curves. This suggests that the key challenge lies in localizing edits while preserving the surrounding relational structure.

\section{Conclusion}
We presented Edit2TikZ, a benchmark for evaluating end-to-end scientific figure editing through compilable TikZ code. Our results show that this task remains difficult even for strong MLLMs, mainly because accurate editing depends on both faithful figure reconstruction and precise localized modification. To better assess this capability, our RS and ECS metrics separately measure non-target preservation and edit correctness, providing a human-aligned evaluation of figure editing quality. TikZEditMix further demonstrates that targeted curriculum training can substantially improve compact models. We hope Edit2TikZ provides a useful foundation for building more reliable systems for scientific figure editing.

\bibliography{aaai2027}

\end{document}